\documentclass[runningheads]{llncs}
\usepackage{makeidx}
\usepackage{graphicx}
\usepackage{amsmath,amssymb} %
\usepackage{color}

\usepackage{subcaption}
\usepackage{mathrsfs}

\newcommand*{\maskname}{central instance~}
\newcommand*{\sparseBr}{sparse-neighborhood~}
\newcommand*{\denseBr}{dense-neighborhood~}
\newcommand*{\encBr}{encoded-neighborhood~}

\newcommand*{\evidW}{w}

\newcommand\coord{\vec}

\usepackage{tabularx}
\usepackage{multirow}
\usepackage{makecell}
\usepackage{textcomp}
\usepackage{booktabs}
\usepackage{marvosym}
\newcolumntype{M}[1]{>{\centering\arraybackslash}m{#1}}
\newcolumntype{R}[1]{>{\raggedleft\arraybackslash}m{#1}} 
\newcolumntype{L}[1]{>{\raggedright\arraybackslash}m{#1}} 
\newcolumntype{?}{!{\vrule width 0.3em}}

\usepackage{algorithm}%
\PassOptionsToPackage{noend}{algpseudocode}%
\usepackage{algpseudocode}%
\algrenewcommand\algorithmicindent{0.8em}

\makeatletter

\renewcommand{\ALG@beginalgorithmic}{\small}
\makeatother

\usepackage{hyperref}

\begin{document}
	\pagestyle{headings}
	\mainmatter

	\def\GCPR20SubNumber{43}

	\title{Proposal-Free Volumetric Instance Segmentation from Latent Single-Instance Masks}

	\titlerunning{Proposal-Free Instance Segmentation from Latent Single-Instance Masks}
	\author{Alberto Bailoni\inst{1}\and
	Constantin Pape\inst{2} \and
	Steffen Wolf\inst{1} \and \\
	Anna Kreshuk\inst{2} \and
	Fred A. Hamprecht\inst{1}}
	\authorrunning{A. Bailoni et al.}
	\institute{HCI/IWR, Heidelberg University, 69120 Heidelberg, Germany 
	\email{\{name.surname\}@iwr.uni-heidelberg.de} \and
	EMBL, 69117 Heidelberg, Germany\\
	\email{\{name.surname\}@embl.de}}

	\maketitle

\begin{abstract}
This work introduces a new proposal-free instance segmentation method that builds on single-instance segmentation masks predicted across the entire image in a sliding window style.
In contrast to related approaches, our method concurrently predicts all masks, one for each pixel, and thus resolves any conflict jointly across the entire image.
Specifically, predictions from overlapping masks are combined into edge weights of a signed graph that is subsequently partitioned to obtain all final instances concurrently.
The result is a parameter-free method that is strongly robust to noise and prioritizes predictions with the highest consensus across overlapping masks. 
All masks are decoded from a low dimensional latent representation, which results in great memory savings strictly required for applications to large volumetric images. 
We test our method on the challenging CREMI 2016 neuron segmentation benchmark where it achieves competitive scores. 
\end{abstract}

	\begin{figure}[t]
\centering
        \includegraphics[width=\textwidth]{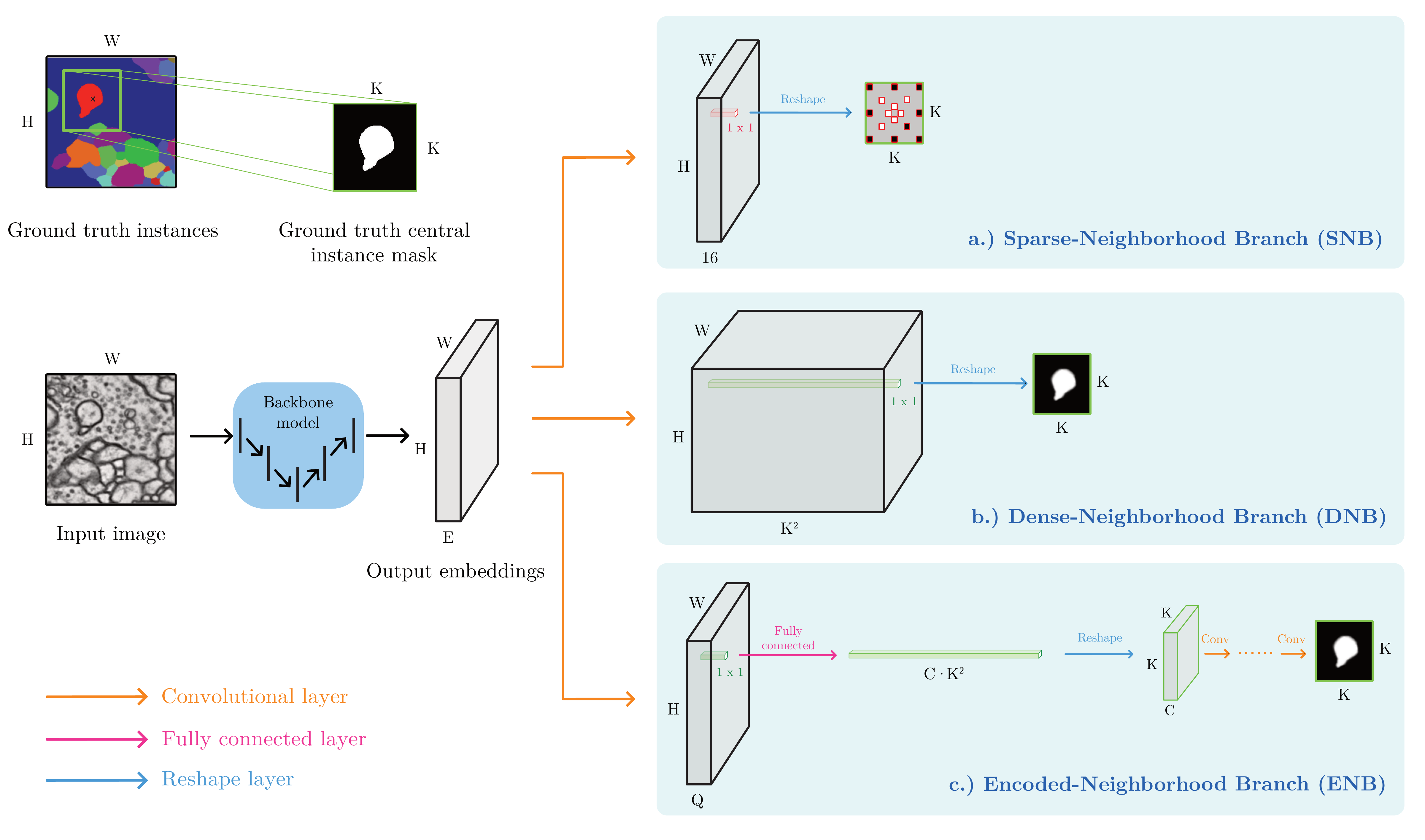} %
        \caption{Comparison between the proposed method and the strong baseline representing the current state-of-the-art. \textbf{Left}: At the top-left corner, an example of binary \maskname mask for a given ground truth label image; below, the backbone model predicts feature maps with the spatial dimensions of the input image. \textbf{Right}: a.) \emph{Sparse-neighborhood branch} used in the baseline model to predict affinities for a given sparse neighborhood structure; b) Simple generalization of the \emph{\sparseBr branch} to predict dense \maskname masks; c) Proposed \emph{\encBr branch} predicting \maskname masks in a low-dimensional latent space.}
    \label{fig:main_figure}
\end{figure}

\section{Introduction}\label{sec:intro}

\emph{Instance segmentation} is the computer vision task of assigning each pixel of an image to an instance, such as individual car, person or biological cell. %
There are two main types of successful deep learning approaches to instance segmentation: \emph{proposal-based} and \emph{proposal-free} methods. 
Recently, there has been a growing interest in the latter. Proposal-free methods do not require object detection and are preferred in imagery as studied here, in which object instances cannot be approximated by bounding boxes and are much larger than the field of view of the model.

Some recent successful proposal-free approaches \cite{januszewski2018high,liu2016multi,meirovitch2016multi} tackle instance segmentation by predicting, for a given patch of the input image, whether or not each pixel in the patch is part of the instance that covers the central pixel of the patch. 
This results a probability mask, which from now on we call \emph{\maskname mask}. These masks are then repeatedly predicted across the entire image, either in a sliding window style or by starting from a seed and then shifting the field of view depending on the previously predicted masks. 
Final object-instances are then obtained by aggregating predictions from overlapping masks which is in itself a nontrivial and interesting problem.

In this work, we introduce a new proposal-free segmentation method that is also based on predicting \maskname masks\footnote{For interesting, closely related but independent work, see \cite{hirsch2020patchperpix}.}.
However, our approach comes with four main advantages compared to previous methods.
Firstly, our model concurrently predicts all \maskname masks, one for each pixel, by using a fully-convolutional approach with much smaller computational footprint than previous methods, which iteratively predict one instance at the time, one mask after the other \cite{januszewski2018high,meirovitch2016multi}.
Secondly, our approach predicts \maskname masks in a low dimensional latent representation (see Fig. \ref{fig:main_figure}c), which results in great memory savings that are strictly required to apply the method to large volumetric images. 
Thirdly, the proposed approach aggregates predictions from overlapping \maskname masks without the need for any extra parameter or threshold and outputs predictions with associated uncertainty;
and, finally, all final object-instances are obtained concurrently, as opposed to previous methods predicting them one-by-one with subsequent conflict resolution.

Additionally, we systematically compare the proposed model with the current state-of-the-art proposal-free method both on natural and biological images \cite{liu2018affinity,Gao_2019_ICCV,lee2017superhuman,wolf2018mutex,bailoni2019generalized}. This strong baseline consists of a fully-convolutional network predicting, for each pixel, an arbitrary predefined set of short- and long-range affinities, i.e. neighborhood relations representing how likely it is for a pair of pixels to belong to the same object instance (see Fig. \ref{fig:main_figure}a). 

Our method achieves competitive scores on the challenging CREMI 2016 neuron segmentation benchmark. In our set of validation experiments, we show how predicting encoded \maskname masks always improves accuracy. Moreover, when predictions from overlapping masks are combined into edge weights of a graph that is subsequently partitioned, the result is a method that is strongly robust to noise and gives priority to predictions sharing the highest consensus across predicted masks. 
This parameter-free algorithm, for the first time, outperforms super-pixel based methods, which have so far been the default choice on the challenging data from the CREMI competition challenge.

\section{Related Work} \label{sec:related_work}
Many of the recent successful instance segmentation methods on natural images are \emph{proposal-based}: they first perform object detection, for example by predicting anchor boxes \cite{ren2015faster}, and then assign a class and a binary segmentation mask to each detected bounding box \cite{he2017mask,porzi2019seamless}.
Proposal-Free methods on the other hand directly group pixels into instances. 
Recent approaches use metric learning to predict high-dimensional associative pixel embeddings that map pixels of the same instance close to each other, while mapping pixels belonging to different instances further apart, e.g. \cite{lee2019learning,kong2018recurrentPix}. %
Final instances are then retrieved by applying a clustering algorithm. A post-processing step is needed to merge instances that are larger then the field of view of the network. 

\textbf{Aggregating Central Instance Masks} -- 
The line of research closest to ours predicts overlapping \maskname masks in a sliding window style across the entire image. The work of \cite{liu2016multi} aggregates overlapping masks and computes intersection over union scores between them.
In neuron segmentation, flood-filling networks \cite{januszewski2018high} and MaskExtend \cite{meirovitch2016multi} use a CNN to iteratively grow one instance/neuron at a time, merging one mask after the other. Recently, the work of \cite{meirovitch2019cross} made the process more efficient by employing a combinatorial encoding of the segmentation, but the method remains orders of magnitude slower as compared to the convolutional one proposed here, since in our case all masks are predicted at the same time and for all instances at once.
The most closely related work to ours is the independent preprint \cite{hirsch2020patchperpix}, where a very similar model is applied to the BBBC010 benchmark microscopy dataset of \emph{C. elegans} worms. However, here we propose a more efficient model that scales to 3D data, and we provide an extensive comparison to related models predicting long-range pixel-pair affinities. 

\textbf{Predicting Pixel-Pair Affinities} --  
Instance-aware edge detection has experienced recent progress due to deep learning, both on natural images and biological data \cite{Gao_2019_ICCV,liu2018affinity,lee2017superhuman,wolf2018mutex,schmidt2018cell,zeng2017deepem3d,parag2017anisotropic,bailoni2019generalized}. Among these methods, the most recent ones also predict long-range affinities between pixels and not only direct-neighbor relationships \cite{Gao_2019_ICCV,liu2018affinity,lee2017superhuman}.
Other related work \cite{funke2018large,turaga2009maximin} approach boundary detection via a structured learning approach.
In neuron segmentation, boundaries predicted by a CNN are converted to final instances with subsequent postprocessing and superpixel-merging.
Some methods define a graph with both positive and negative weights and formulate the problem in a combinatorial framework, known as multicut or correlation clustering problem \cite{chopra1991multiway}. 
In neuron segmentation and connectomics, exact solvers can tackle problems of considerable size \cite{andres2012globally}, but accurate approximations \cite{pape2017solving,yarkony2012fast} and greedy agglomerative algorithms \cite{levinkov2017comparative,wolf2019mutex,bailoni2019generalized} are required on larger problems.

\section{Model and Training Strategy}\label{sec:model}
In this section, we first define \maskname masks in Sec.~\ref{sec:self_masks}.
Then, in Sec. \ref{sec:encoding_masks}, we present our first main contribution, a model trained end-to-end to predict encoded \maskname masks, one for each pixel of the input image. 

\subsection{Local Central Instance Masks}\label{sec:self_masks}
This work proposes to distinguish between different object instances based on instance-aware pixel-pair affinities in the interval $[0,1]$, which specify whether or not two pixels belong to the same instance or not.
Given a pixel of the input image with coordinates $\coord{u}= (u_x, u_y)$, a set of affinities to neighboring pixels within a $K\times K$ window is learned, where $K$ is an odd number. 
We define the $K\times K$-neighborhood of a pixel as: 
$\mathcal{N}_{K\times K} \equiv \mathcal{N}_{K} \times \mathcal{N}_{K}$, where $\mathcal{N}_{K} \equiv \left\{-\frac{K-1}{2}, \ldots, \frac{K-1}{2}\right\}$ and represent the affinities relative to pixel $\coord{u}$ as a \maskname mask $\mathcal{M}_{\coord{u}}: \mathcal{N}_{K\times K} \rightarrow [0,1]$.

We represent the associated training targets as binary ground-truth masks $\hat{\mathcal{M}}_{\coord{u}}: \mathcal{N}_{K\times K} \rightarrow \{0,1\}$, which can be derived from a ground-truth instance label image $\hat{L}: H\times W \rightarrow \mathbb{N}$ with dimension $H\times W$:
\begin{equation}\label{eq:target_masks}
\forall\, \coord{u}\in H\times W, \quad \forall\, \coord{n}\in \mathcal{N}_{K\times K} \qquad \hat{\mathcal{M}}_{\coord{u}}(\coord{n}) = 
\begin{cases}
1, \quad &\text{if } \hat{L}(\coord{u}) = \hat{L}(\coord{u}+\coord{n}) \\
0, \quad & \text{otherwise}.
\end{cases}
\end{equation}
We actually use similar definitions in 3D, but use 2D notation here for simplicity.

\begin{figure}[t]
\centering
\begin{minipage}[t]{0.36\textwidth}
        \centering
        \includegraphics[width=0.99\textwidth]{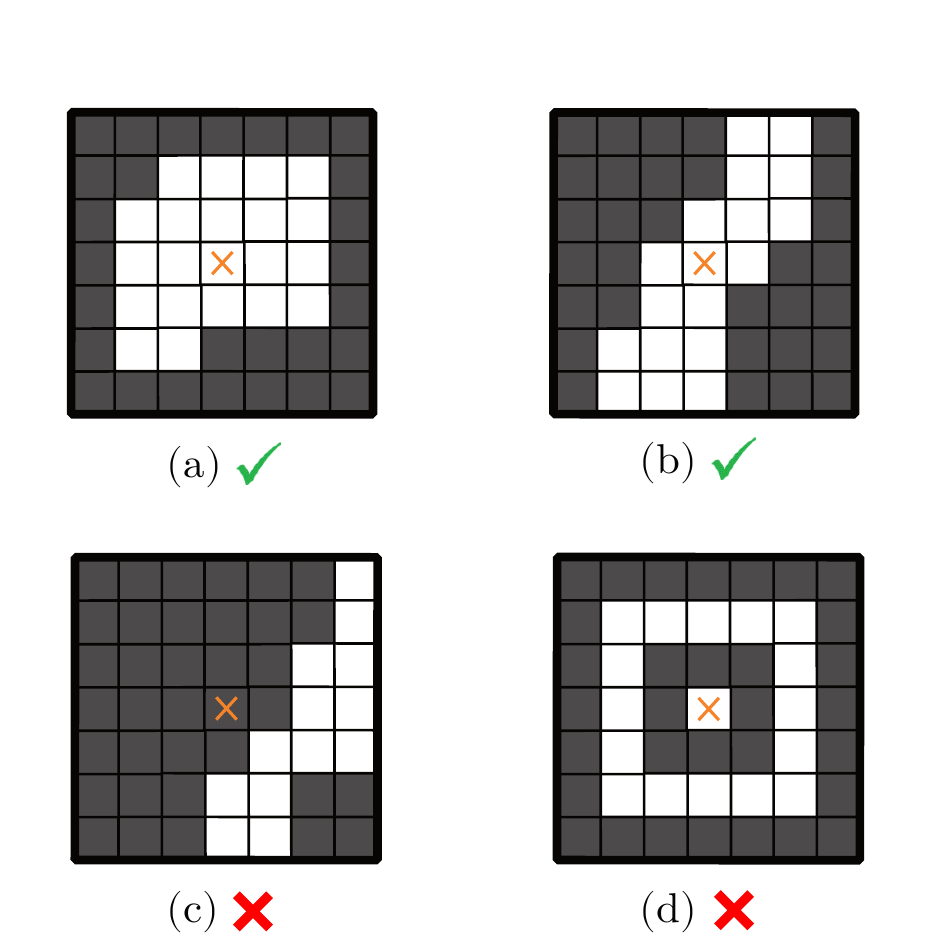} %
        \captionof{figure}{Examples of expected (\textbf{a-b}) and not expected (\textbf{c-d}) binary 2D \maskname masks. }
    \label{fig:valid_masks}
\end{minipage}\hfill
\begin{minipage}[t]{0.58\textwidth}
\centering
\includegraphics[width=0.99\textwidth]{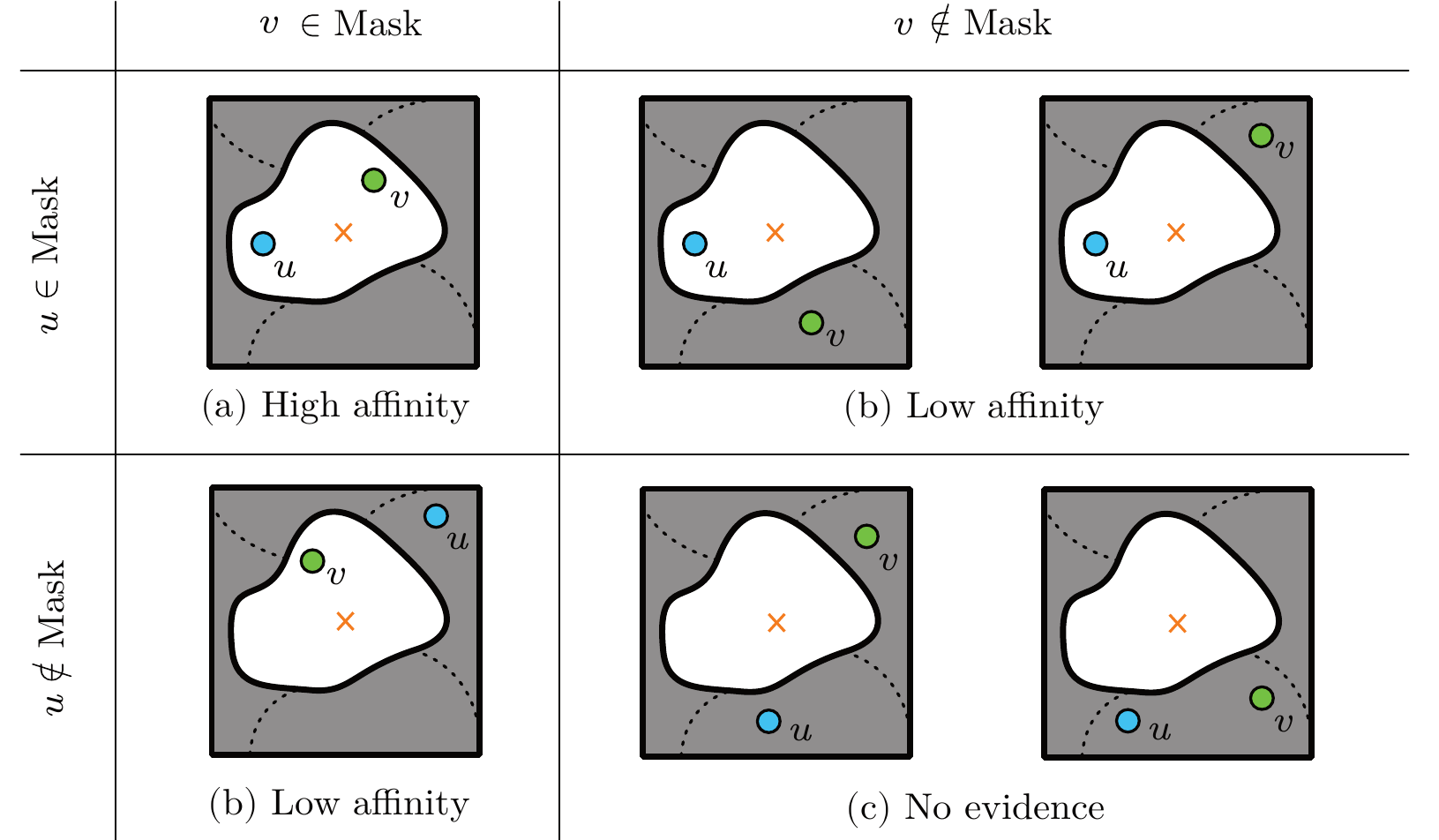} %
        \captionof{figure}{Computing instance-aware affinity between pixels $u$ and $v$ from instance masks associated to the central pixel in the patch (orange cross). 
        }
    \label{fig:mask_cases}
\end{minipage}
\end{figure}

\subsection{Training Encoded Central Instance Masks End-To-End}\label{sec:encoding_masks}
In several related work approaches \cite{liu2018affinity,Gao_2019_ICCV,lee2017superhuman,wolf2018mutex,bailoni2019generalized}, affinities between pairs of pixels are predicted for a predefined sparse stencil representing a set of $N$ short- and long-range neighborhood relations for each pixel  ($N=8$ \emph{\sparseBr branch} of Fig. \ref{fig:main_figure}a). The $N$ output feature maps  are then trained with a binary classification loss.

On paper, this training method can be easily generalized to output a feature map of size $K^2 \times H \times W$ and thus predict a full $K\times K$ \maskname mask for each pixel of the input image (see \emph{\denseBr branch} in Fig. \ref{fig:main_figure}b).
Nevertheless, in practice, this model has prohibitively large memory requirements for meaningful values of $K$, precluding application to 3D data of interest here.

However, among the $2^{K\cdot K}$ conceivable binary masks $\hat{\mathcal{M}}_{\coord{u}}: \mathcal{N}_{K^2} \rightarrow \{0,1\}$, in practice only a tiny fraction corresponds to meaningful instance masks (see some examples in Fig. \ref{fig:valid_masks}). 
This suggests that it is possible to find a compact representation that spans the manifold of expected instance shapes.

As our first main contribution, we test this assumption by training a model end-to-end to predict, for each pixel $\coord{u}\in H\times W$ of the input image, a latent vector $z_{\coord{u}}\in \mathbb{R}^Q$ encoding the $K \times K$ \maskname mask $\mathcal{M}_{\coord{u}}$ centered at pixel $\coord{u}$ (see \emph{\encBr branch} in Fig. \ref{fig:main_figure}c). 
The backbone model is first trained to output a more compact $Q\times H\times W$ feature map and
then a tiny convolutional decoder network is applied to each pixel of the feature map to decode masks.
During training, decoding one mask for each pixel in the image would be too memory consuming. Thus, we randomly sample $R$ pixels with coordinates $\coord{u}_1, \ldots, \coord{u}_R$ and only decode the associated masks $\mathcal{M}_{\coord{u}_1}, \ldots, \mathcal{M}_{\coord{u}_R}$. 
Given the ground-truth \maskname masks $\hat{\mathcal{M}}_{\coord{u}_i}$ defined in Eq. \ref{eq:target_masks}, the training loss is then defined according to the S\o rensen-Dice coefficient formulated for fuzzy set membership values, similarly to what was done in \cite{wolf2018mutex}.
Ground-truth labels are not always pixel-precise and it is often impossible to estimate the correct label for pixels that are close to a ground-truth label transition. Thus, in order to avoid noise during training, we predict completely empty masks for pixels that are less than two pixels away from a label transition, so that the model is trained to predict single-pixel clusters along the ground-truth boundaries. In our experiments, this approach performed better than masking the training loss along the boundaries.

\subsection{Predicting Multi-Scale Central Instance Masks}\label{sec:multiscale_patches}
Previous related work \cite{lee2017superhuman,liu2018affinity,Gao_2019_ICCV} shows that predicting long-range affinities between distant pixels improves accuracy as compared to predicting only short-range ones. However, predicting large \maskname masks would translate to a bigger model that, on 3D data, would have to be trained on a small 3D input field of view.
This, in practice, usually decreases accuracy because of the reduced 3D context available to the network.
Thus, we instead predict multiple \maskname masks of the same window size $7 \times 7 \times 5$ but at different resolutions, so that the lower the resolution the larger the size of the associated patch in the input image. 
These multiple masks at different resolutions are predicted by adding several \emph{\encBr branches} along the hierarchy of the decoder in the backbone model, which in our case is a 3D U-Net \cite{ronneberger2015u,cciccek20163d} (see Fig.~\ref{fig:model_architecture}). 
In this way, the encoded \maskname masks at higher and lower resolutions can be effectively learned at different levels in the feature pyramid of the U-Net.

\begin{algorithm}[t]
  \begin{flushleft}
  \caption{: Affinities from Aggregated Central Instance Masks}
   \hspace*{\algorithmicindent} \textbf{Input:} Graph $\mathcal{G}(V,E)$; \maskname masks $\mathcal{M}_{\coord{u}}: \mathcal{N}_{K\times K} \rightarrow [0,1]$  \\
  \hspace*{\algorithmicindent} \textbf{Output:} Affinities $\bar{a}_e\in[0,1]$ with variance $\sigma^2_e$ for all edges $e\in E$\\
  \hspace*{\algorithmicindent} 
  \begin{algorithmic}[1]
  \footnotesize
      \For{each edge $e=(\coord{u}, \coord{v})\in E$ in graph $\mathcal{G}$}
        \State Get coordinates $\coord{u}=(u_x,u_y)$ and $\coord{v}=(v_x,v_y)$ of pixels linked by edge $e$
        \State Collect all $T$ masks $\mathcal{M}_{\coord{c}_1},\ldots,\mathcal{M}_{\coord{c}_T}$ including both pixel $\coord{u}$ and pixel $\coord{v}$
        \State Init. vectors $[a_1,\ldots,a_T] = [\evidW_1,\ldots,\evidW_T] = 0$ for affinities and evidence weights
        \For{$i\in 1, \ldots,T$}
            \State Get relative coords. of $\coord{u}$ and $\coord{v}$ with respect to the central pixel $\coord{c}_i$
            \State $a_i \gets \min \big(\mathcal{M}_{\coord{c}_i}(\coord{u} - \coord{c}_i), \,\mathcal{M}_{\coord{c}_i}(\coord{v} - \coord{c}_i)\big)$ \Comment{Fuzzy-AND: both values active}
            \State $\evidW_i \gets \max \big(\mathcal{M}_{\coord{c}_i}(\coord{u} - \coord{c}_i), \,\mathcal{M}_{\coord{c}_i}(\coord{v} - \coord{c}_i)\big)$ \Comment{Fuzzy-OR: at least one value active}
        \EndFor
        \State Get weighted affinity average $\bar{a}_e= \sum_{i} a_i \evidW_i\,/\,\sum_{i}\evidW_i$ 
        \State Get weighted affinity variance $\sigma^2_e = \sum_{i} \evidW_i (a_i-\bar{a}_e)^2\,/\,\sum_{i}\evidW_i$
      \EndFor
      \State
      \Return $\bar{a}_e, \sigma^2_e$ for each $e\in E$
  \end{algorithmic}
    \label{alg:computing_affinities}
  \end{flushleft}

\end{algorithm}
\begin{figure}[t]
\centering
        \includegraphics[width=0.9\textwidth]{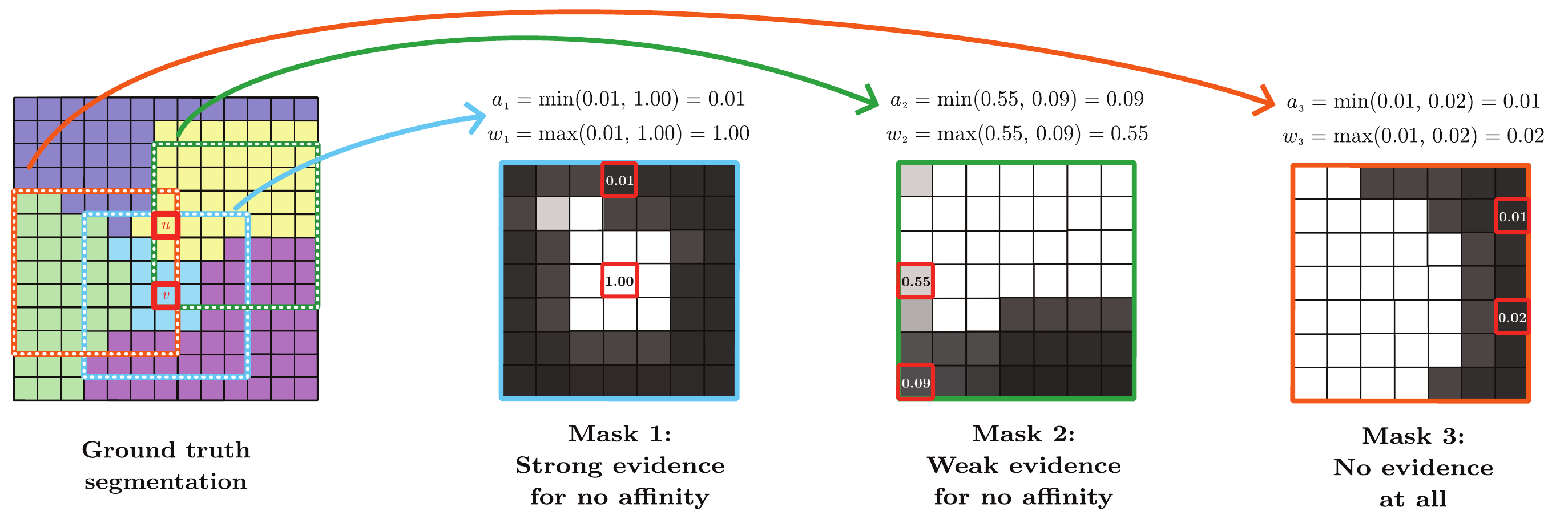} %
        \caption{Proposed method to average overlapping masks and compute the affinity between pixel $u$ and pixel $v$ (highlighted in red in the ground-truth segmentation on the left). For simplicity, we only consider three masks among all the ones including both pixels $u$ and $v$. 
        In \emph{Mask 1}, only $v$ is part of the mask, so there is a strong evidence for no affinity between $u$ and $v$; in \emph{Mask 2},  $u$ is predicted to be part of the mask only with a low confidence, so the contribution of this mask in the final average will be weak; in \emph{Mask 3}, both pixels are not part of the \maskname mask, so there is no evidence about their affinity. 
        The final affinity value of edge $(u,v)$ is given by the weighted average of the collected affinities $a_i$ weighted with the evidence weights $\evidW_i$: $\bar{a}_e= \sum_{i=1}^3 a_i \evidW_i\,/\,\sum_{i}\evidW_i$
        }
    \label{fig:alg_explained}
\end{figure}

\section{Affinities with Uncertainty from Aggregated Masks}\label{sec:aggr_affs}
In order to obtain an instance segmentation from the predictions of the model presented in Sec. \ref{sec:model}, we now compute instance-aware pixel-pair affinities for a given sparse $N$-neighborhood structure (see Table \ref{tab:neighborhood_structures} in supplementary material for details about the structure) and use them as edge weights of a pixel grid-graph $\mathcal{G}(V,E)$, such that each node represents a pixel / voxel of the image. The graph is then partitioned to obtain object instances.

In this section, we propose an algorithm that, without the need of any threshold parameter, aggregates predictions from overlapping \maskname masks and outputs edge weights with associated uncertainty.
Related work either thresholds the predicted \maskname masks \cite{januszewski2018high,hirsch2020patchperpix,meirovitch2016multi} or computes Intersection over Union (IoU) scores for overlapping patches \cite{liu2016multi}. However, an advantage of predicting pixel-pair affinities / pseudo-probabilities as compared to IoU scores is that affinities can easily be translated into attractive and repulsive interactions in the grid-graph 
and a parameter-free partitioning algorithm can be employed to yield instances.

Here, we propose a simple algorithm to aggregate predictions from multiple patches: Fig. \ref{fig:alg_explained} shows a simplified example of how Algorithm \ref{alg:computing_affinities} computes the affinity for an edge $e$ linking a pair of pixels $\coord{u}$ and $\coord{v}$.
As a first step, the algorithm loops over all predicted \maskname masks including both $\coord{u}$ and $\coord{v}$. 
However, not all these masks are informative, as we visually explain in Fig.~\ref{fig:mask_cases}: a mask $\mathcal{M}_{\coord{c}_i}$ centered at pixel $\coord{c}_i$ provides any evidence about the affinity between pixels $\coord{u}$ and $\coord{v}$ only if at least one of the two pixels belongs to the mask (fuzzy OR operator at line 8 in Alg. \ref{alg:computing_affinities}).
If both pixels do not belong to it, we cannot say anything about whether they belong to the same instance (see Fig. \ref{fig:mask_cases}c). We model this with an evidence weight $\evidW_i\in[0,1]$, which is low when both pixels do not belong to the mask.
On the other hand, when at least one of the two pixels belongs to the mask, we distinguish two cases (fuzzy AND operator at line 7 in Alg. \ref{alg:computing_affinities}): i)
both pixels belong to the mask  (case in Fig. \ref{fig:mask_cases}a), so by transitivity we conclude they should be in the same instance and their affinity $a_i$ should tend to one; 
ii) only one pixel belongs to the mask (case in Fig. \ref{fig:mask_cases}b), so that according to this mask they are in different instances and their affinity should tend to zero. 

At the end, we compute a weighted average $\bar{a}_e$ and variance $\sigma^2_e$ of the collected affinities from all overlapping masks, such that masks with more evidence will contribute more on average, and the obtained variance is a measure of how consistent were the predictions across masks. 
The algorithm was implemented on GPU using PyTorch and the variance was computed via Welford's online stable algorithm \cite{welford1962note}.

\section{Experiments on Neuron Segmentation}
We evaluate and compare our method on the task of neuron segmentation in electron microscopy (EM) image volumes. This application is of key interest in connectomics, a field of neuro-science with the goal of reconstructing neural wiring diagrams spanning complete central nervous systems. 
We test our method on the competitive CREMI 2016 EM Segmentation Challenge \cite{cremiChallenge}. 
We use the second half of CREMI sample C as validation set for our comparison experiments in Table \ref{tab:val_results} and then we train a final model on all the three samples with available ground truth labels to submit results to the leader-board in Tab. \ref{tab:test_results}. 
Results  are evaluated using the CREMI score, which is given by the geometric mean of Variation of Information Score (VOI split + VOI merge) and Adapted Rand-Score (Rand-Score) \cite{arganda2015crowdsourcing}. See Sec.~\ref{sec:cremi_data_augm} for more details on data augmentation, strongly inspired by related work.

\subsection{Architecture details of the tested models}\label{sec:models_details}
As a backbone model we use a 3D U-Net consisting of a hierarchy of four feature maps with anisotropic downscaling factors $(\frac{1}{2},\frac{1}{2},1)$, similarly to \cite{lee2019learning,lee2017superhuman,wolf2018mutex}. 
Models are trained with the Adam optimizer and a batch size equal to one. Before applying the loss, we slightly crop the predictions to prevent training on borders where not enough surrounding context is provided. 
See Sec. \ref{sec:arch_details_suppl} and Fig. \ref{fig:model_architecture} in supplementary material for all details about the used architecture.

\textbf{Baseline Model (SNB)} -- As a strong baseline, we re-implement the current state-of-the-art and train a model to predict affinities for a sparse neighborhood structure (Fig. \ref{fig:main_figure}a). We perform deep supervision by attaching three \emph{\sparseBr branches} (SNB) at different levels in the hierarchy of the UNet decoder and train the coarser feature maps to predict longer range affinities. Details about the used neighborhood structures and the architecture can be found in Table \ref{tab:neighborhood_structures} and Fig. \ref{fig:model_architecture}.

\textbf{Proposed Model (ENB)} -- We then train a model to predict encoded \maskname masks (Fig. \ref{fig:main_figure}c). Similarly to the baseline model, we provide deep supervision by attaching four \emph{\encBr branches} (ENB) to the backbone U-Net. As explained in Sec. \ref{sec:multiscale_patches}, all branches predict 3D masks of shape $7 \times 7 \times 5$, but at different resolutions $(1,1,1)$, $(\frac{1}{4},\frac{1}{4},1)$ and $(\frac{1}{8},\frac{1}{8},1)$, as we show in the architecture in Fig. \ref{fig:model_architecture}. A visualization of the learned latent spaces is given in Fig. \ref{fig:PCA_embeddings}.

\textbf{Combined Model (SNB+ENB)} -- Finally, we also train a combined model to predict both \maskname masks and a sparse neighborhood of affinities, by providing deep supervision both via \emph{\encBr} and \emph{\sparseBr} \emph{branches}. The backbone of this model is then trained with a total of seven branches: three branches equivalent to the ones used in the baseline model SNB, plus four additional ones like those in the ENB model (see Fig. \ref{fig:model_architecture}).  

\subsection{Graph Partitioning Methods} 
Given the predicted encoded \maskname masks, we compute affinities $a_e$ either with the average aggregation method introduced in Sec. \ref{sec:aggr_affs} (\textbf{MaskAggr}) or the efficient approach described in Sec. \ref{sec:efficient_affs}. 
The result of either is a signed pixel grid-graph, i.e. a graph with positive and negative edge weights that needs to be partitioned into instances. 
The used neighborhood connectivity of the graph is given in Table \ref{tab:neighborhood_structures}. Positive and negative edge weights $w_e$ are computed by applying the additive transformation $w_e=a_e-0.5$ to the predicted affinities.

To obtain final instances, we test different partitioning algorithms.
The Mutex Watershed (\textbf{MWS}) \cite{wolf2018mutex} is an efficient algorithm to partition graphs with both attractive and repulsive weights without the need for extra parameters. It can easily handle the large graphs considered here with up to $10^8$ nodes/voxels and $10^9$ edges\footnote{Among all edges given by the chosen neighborhood structure, we add only 10\% of the long-range ones, since the Mutex Watershed was shown to perform optimally in this setup \cite{bailoni2019generalized,wolf2018mutex}.}. 

Then, we also test another graph partitioning pipeline that has often been applied to neuron segmentation because of its robustness. This method first generates a 2D super-pixel over-segmentation from the model predictions and then partitions the associated region-adjacency graph to obtain final instances. Super-pixels are computed with the following procedure: First, the predicted direct-neighbor affinities are averaged over the two isotropic directions to obtain a 2D neuron-membrane probability map; then, for each single 2D image in the stack, super-pixels are generated by running a watershed algorithm seeded at the maxima of the boundary-map distance transform (\textbf{WSDT}). Given this initial over-segmentation, a 3D region-adjacency graph is built, so that each super-pixel is represented by a node in the graph. Edge weights of this graph are computed by averaging short- and long-range affinities over the boundaries of neighboring super-pixels. 
Finally, the graph is partitioned by applying the average agglomeration algorithm proposed in \cite{bailoni2019generalized} (\textbf{GaspAvg}).

\begin{figure}[t]
\begin{subfigure}[t]{0.32\linewidth}
\centering
\includegraphics[width=0.75\linewidth,trim=0in 0in 0in 0.2in,clip]{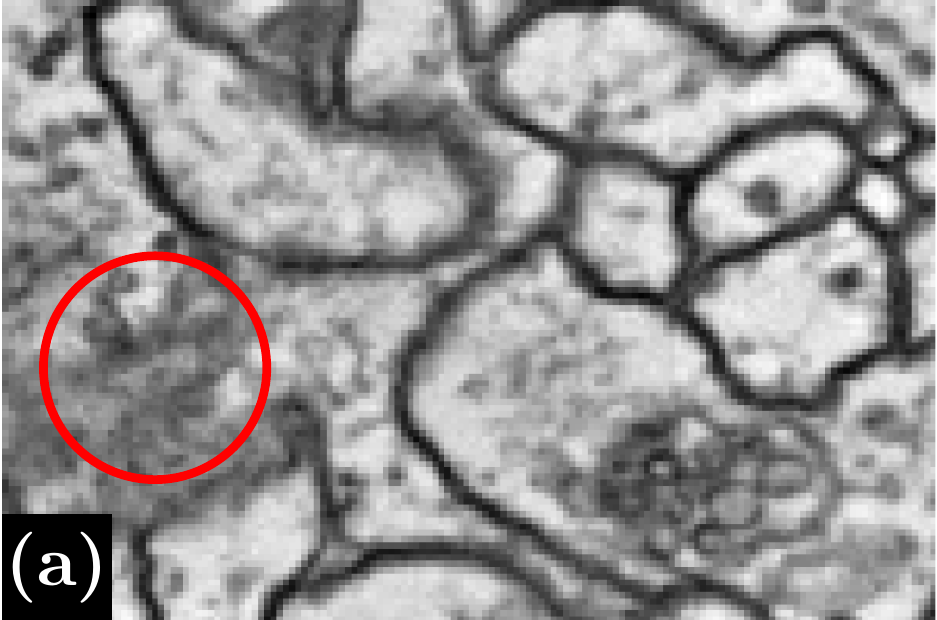} %
\end{subfigure}\hfill
\begin{subfigure}[t]{0.32\textwidth}
\centering
\includegraphics[width=0.75\linewidth,trim=0in 0in 0in 0.2in,clip]{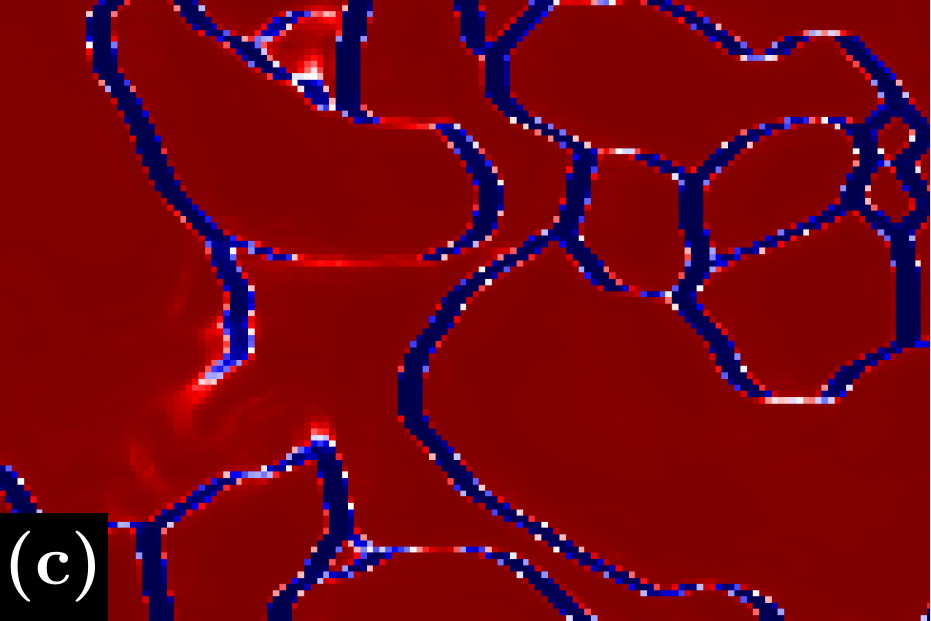} %
\end{subfigure}\hfill
\begin{subfigure}[t]{0.32\linewidth}
\centering
\includegraphics[width=0.75\linewidth,trim=0in 0in 0in 0.2in,clip]{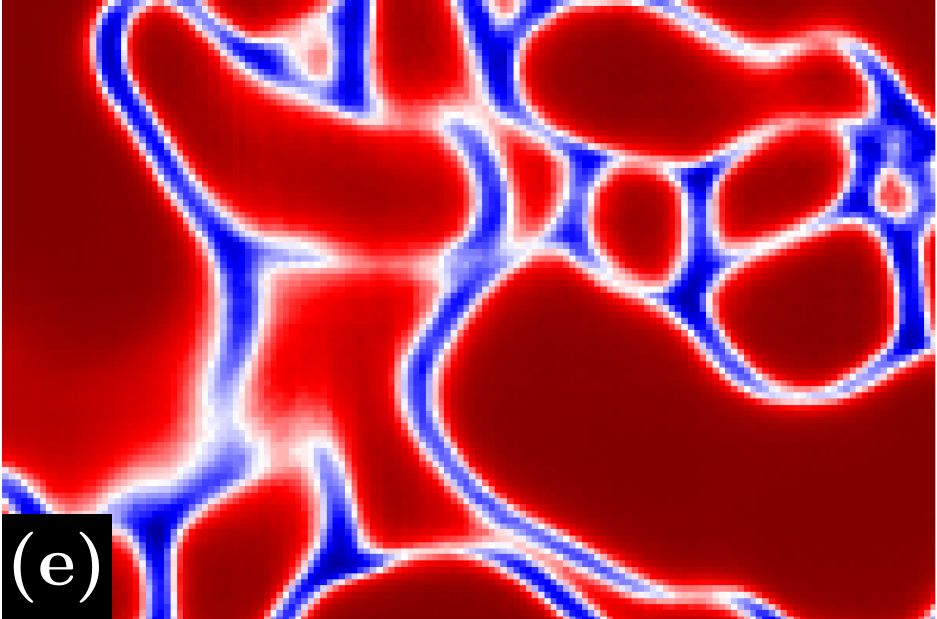} %
\end{subfigure}\vspace{0.6em}\\
\begin{subfigure}[t]{0.32\textwidth}
\centering
\includegraphics[width=0.75\linewidth,trim=0in 0in 0in 0.2in,clip]{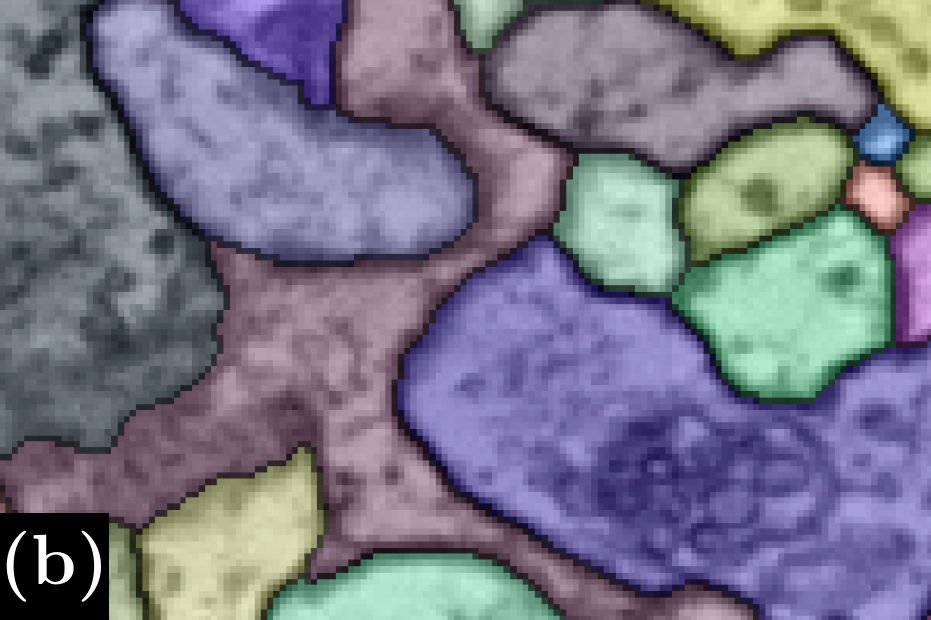} %
\end{subfigure}\hfill
\begin{subfigure}[t]{0.32\linewidth}
\centering
\includegraphics[width=0.75\linewidth,trim=0in 0in 0in 0.2in,clip]{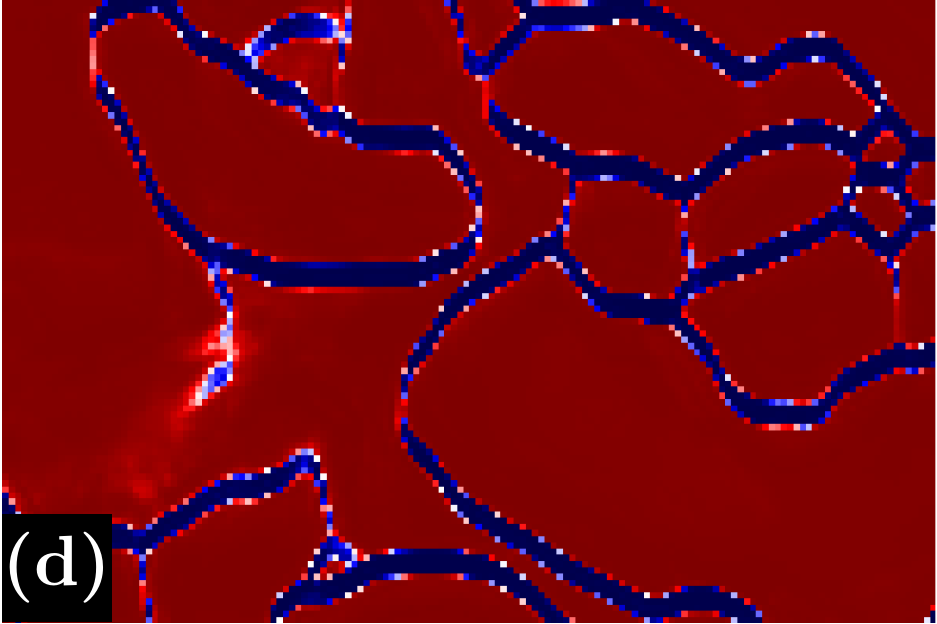} %
\end{subfigure}\hfill
\begin{subfigure}[t]{0.32\textwidth}
\centering
\includegraphics[width=0.75\linewidth,trim=0in 0in 0in 0.2in,clip]{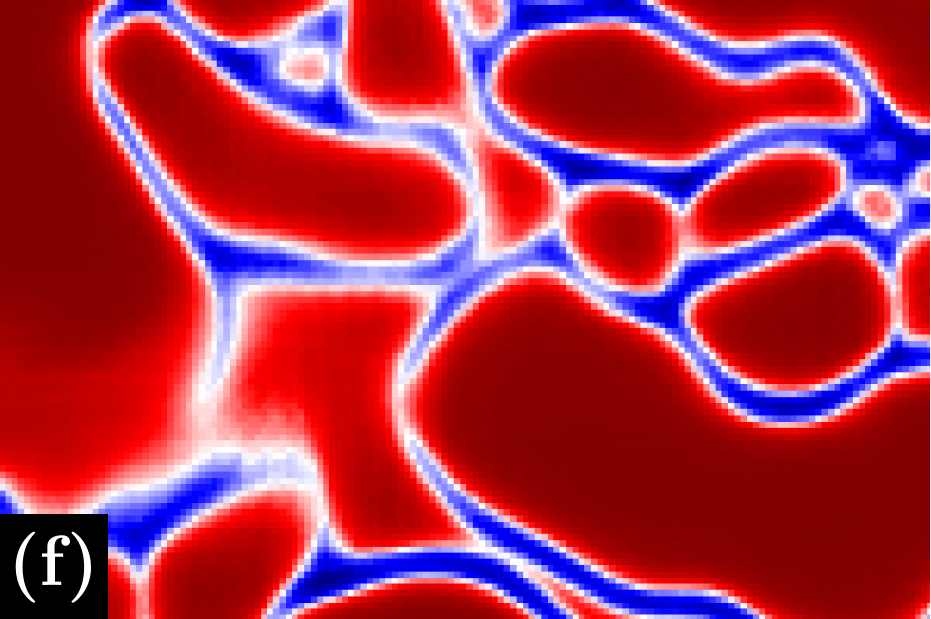} %
\end{subfigure}
\caption{Comparison between different affinities and their robustness to noise. \textbf{\mbox{(a-b)}} Raw data and ground-truth labels. \textbf{(c-d)} Affinities predicted by the \emph{sparse-neighborhood branch}, which is trained with a dense binary classification loss (high affinities are red). \textbf{(e-f)} Affinities computed by averaging overlapping masks as explained in Sec. \ref{sec:aggr_affs} (MaskAggr). Affinities from averaged masks are smoother and present a more consistent boundary evidence in the noisy region highlighted by the red circle in \textbf{(a)}. Here we show affinities along the horizontal (-4, 0, 0) and vertical (0, -4, 0) directions.}\label{fig:affs_comparison}
\end{figure}

\subsection{Results and Discussion}

\textbf{Pre-Training of the Encoded Space} -- The proposed model based on an \emph{\encBr branch} can be properly trained only if the dimension $Q$ of the latent space is large enough to accommodate all possible occurring neighborhood patterns. 
To find a small but sufficiently large $Q$, we trained a convolutional Variational Auto-encoder (VAE) \cite{kingma2013auto,rezende2014stochastic} to compress binary ground-truth \maskname masks $\hat{\mathcal{M}}_{\coord{u}}$ into latent variables $z_{\coord{u}}\in \mathbb{R}^Q$ and evaluated the quality of the reconstructed binary masks via the reconstruction loss. We concluded that $Q=32$ is large enough to compress the masks considered here consisting of $7\times 7 \times 5=245$ pixels. 
As a first experiment, we tried to make use of this VAE-pretrained latent space to train the proposed \emph{\encBr branch} and predict encoded masks directly in this space by using an L2 loss on the encoded vectors. However, similarly to the findings of \cite{hirsch2020patchperpix}, this approach performed worse than directly training the full model end-to-end as described in Sec. \ref{sec:encoding_masks}.

\textbf{Training Encoded Masks} -- As we show in our validation experiments in Tab. \ref{tab:val_results}, models trained to predict encoded \maskname masks (ENB) achieved better scores than the current state-of-the-art method predicting affinities for a sparse neighborhood structure (SNB). 
Our interpretation of this result is that using the encoding process to predict \maskname masks encourages the model to predict segment shapes that are consistent in a larger neighborhood, which can be helpful to correctly segment the most difficult regions of the data. 

\begin{table}[t]
\centering
\scriptsize
        \begin{tabular}[t]{c c c c c c c}
\makecell{Train \\ Sparse \\Neighbor.\\(SNB)} & \makecell{Train\\ Encoded\\Masks\\(ENB)} & \makecell{Aggregate\\Overlapping\\Masks \\(MaskAggr)} & \makecell{Partitioning \\algorithm} &\makecell{No\\superpixels\\required}  & \makecell{CREMI-Score \\(lower is better)} & \makecell{VI-merge \\(lower is better)} \\ \toprule 

\CrossedBox & \CrossedBox & \CrossedBox & MWS & \CrossedBox & \textbf{0.153} & 0.272 \\
\HollowBox & \CrossedBox & \CrossedBox & MWS & \CrossedBox & 0.184 & 0.273 \\
\HollowBox & \CrossedBox & \HollowBox & MWS & \CrossedBox & 0.419 & 0.302 \\
\CrossedBox & \CrossedBox & \HollowBox & MWS & \CrossedBox & 0.532 & 0.447 \\
\CrossedBox & \HollowBox & \HollowBox & MWS & \CrossedBox & 1.155 & 0.874 \\ \midrule
\HollowBox & \CrossedBox & \HollowBox & WSDT+GaspAvg & \HollowBox & 0.173 & \textbf{0.234} \\
\CrossedBox & \CrossedBox & \HollowBox & WSDT+GaspAvg & \HollowBox & 0.237 & 0.331 \\
\CrossedBox & \HollowBox & \HollowBox & WSDT+GaspAvg & \HollowBox & 0.254 & 0.355 \\
\CrossedBox & \CrossedBox & \CrossedBox & WSDT+GaspAvg & \HollowBox& 0.334 & 0.388 \\
\HollowBox & \CrossedBox & \CrossedBox & WSDT+GaspAvg & \HollowBox & 0.357 & 0.391 \\

        \end{tabular}
        \vspace{1em}
        \caption{Comparison experiments on our CREMI validation set. Training encoded \maskname masks (ENB) achieved better scores than the current state-of-the-art approach training only affinities for a sparse neighborhood (SNB). The model that performed best was the one using the method proposed in Sec. \ref{sec:aggr_affs} to average overlapping masks (MaskAggr).} \label{tab:val_results}
\end{table}

\textbf{Aggregating Overlapping Masks}  -- 
In our validation experiments of Tab. \ref{tab:val_results}, we also test the affinities computed by averaging over overlapping masks (MaskAggr), as described in Sec. \ref{sec:aggr_affs}. We then partition the resulting signed graph by using the Mutex Watershed, which has empirical linearithmic complexity in the number of edges. 
Our experiments show that, for the first time on this type of more challenging neuron segmentation data, the Mutex Watershed (MWS) achieves better scores than the super-pixel-based methods (WSDT+GaspAvg), which have so far been known to be more robust to noise but also require the user to tune more hyper-parameters.   

We also note that the MWS achieves competitive scores only with affinities computed from aggregating overlapping masks (MaskAggr).
This shows that the MWS algorithm can take full advantage of the \maskname aggregation process by assigning the highest priority to the edges with largest attractive and repulsive weights that were consistently predicted across overlapping masks.

On the other hand, most of the affinities trained with the \emph{\sparseBr} \linebreak \emph{branch} and a dense binary classification loss are almost binary, i.e. they present values either really close to zero or really close one (see comparison between different types of affinities in Fig. \ref{fig:affs_comparison}).
This is not an ideal setup for the MWS, which is a greedy algorithm merging and constraining clusters according to the most attractive and repulsive weights in the graph.
In fact, in this setting the MWS can often lead to over-segmentation and under-segmentation artifacts like those observed in the output segmentations of the (SNB+ENB+MWS) and (SNB+MWS) models. Common causes of these mistakes can be for example inconsistent predictions from the model and partially missing boundary evidence, which are very common in this type of challenging application (see Fig. \ref{fig:affs_comparison} for an example). 

Finally, we also note that superpixel-based methods did not perform equally well on affinities computed from aggregated masks and the reason is that these methods were particularly tailored to perform well with the more \emph{binary-like} classification output of the \emph{\sparseBr branch}.

\begin{table}[t]
\centering
\scriptsize
\begin{minipage}[t]{\textwidth}
    \centering
        \begin{tabular}[t]{l c c c c c c}
        Model & \makecell{Train \\ Sparse \\Neighbor.\\(SNB)} & \makecell{Train\\ Encoded\\Masks\\(ENB)} & \makecell{Aggregate\\Overlapping\\Masks \\(MaskAggr)} & \makecell{Partitioning\\algorithm} & \makecell{No\\superpixels\\ required}  & \makecell{CREMI-Score \\(lower is better)}  \\ \midrule
GaspUNet\cite{bailoni2019generalized} & \CrossedBox & \HollowBox & \HollowBox & WSDT+LMulticut & \HollowBox & \textbf{0.221} \\
PNIUNet\cite{lee2017superhuman} & \CrossedBox & \HollowBox & \HollowBox & Z-Watershed+Agglo & \HollowBox & 0.228 \\
GaspUNet\cite{bailoni2019generalized} & \CrossedBox & \HollowBox & \HollowBox & GaspAvg & \CrossedBox & 0.241  \\
\textbf{OurUNet} & \CrossedBox & \CrossedBox & \CrossedBox &MWS & \CrossedBox & 0.246  \\
\textbf{OurUNet} & \HollowBox & \CrossedBox & \HollowBox &  WSDT+GaspAvg  & \HollowBox & 0.268  \\
MALAUNet\cite{funke2018large} & \CrossedBox & \HollowBox & \HollowBox & WSDT+Multicut & \HollowBox & 0.276  \\
\textbf{OurUNet} & \CrossedBox & \CrossedBox & \HollowBox & WSDT+GaspAvg & \HollowBox & 0.280  \\
CRUNet\cite{zeng2017deepem3d} & \HollowBox & \HollowBox & \HollowBox & 3D-Watershed & \HollowBox & 0.566   \\
LFC\cite{parag2017anisotropic} & \CrossedBox & \HollowBox & \HollowBox & Z-Watershed+Agglo & \HollowBox & 0.616  \\
        \end{tabular}
        \vspace*{0.99em}
    \caption{Representative excerpt of the published methods currently part of the CREMI leaderboard \cite{cremiChallenge} (July 2020). The best method proposed in this work achieves competitive scores and is based on an efficient parameter-free algorithm that does not rely on superpixels. For more details about the partitioning algorithms used by related work, see references in the first column.}
    \label{tab:test_results}
\end{minipage}
\end{table}

\textbf{Training Both Masks and a Sparse Neighborhood} -- In our validation experiments, the combined model, which was trained to predict both a sparse neighborhood (SNB) and encoded \maskname masks (ENB), achieved the best scores and yielded sharper and more accurate mask predictions.
In general, providing losses for multiple tasks simultaneously has often been proven beneficial in a supervised learning setting.
Moreover, the dense gradient of the \emph{\encBr branch}, which focuses on locally correct predictions, nicely complements the sparse gradient\footnote{The gradient of the \emph{\encBr branch} is sparse, due to GPU-memory constraints as explained in Sec. \ref{sec:encoding_masks}.} of the \emph{\encBr branch}, which focuses on predictions that are consistent in a larger neighborhood. We expect this to be another reason why the combination of affinities and \maskname masks performed best in our experiments.

\textbf{Results on Test Samples} -- The evaluation on the three test samples presented in Tab. \ref{tab:test_results} confirms our findings from the validation experiments: among the methods tested in this work, the best scores are achieved by the combined model (ENB+SNB) and by using the Mutex Watershed algorithm (MWS) on affinities averaged over overlapping masks (MaskAggr).
Our method achieves comparable scores to the only other method in the leader-board that does not rely on super-pixels (line 3 in Table \ref{tab:test_results}). This method uses the average agglomeration algorithm GaspAvg proposed in \cite{bailoni2019generalized} instead of the MWS. GaspAvg has been shown to be more robust to noise than Mutex Watershed, however it is also considerably more computationally expensive to run on large graphs like the ones considered here.

\section{Conclusions}
We have presented a new proposal-free method predicting encoded \maskname masks in a sliding window style, one for each pixel of the input image, and introduced a parameter-free approach to aggregate predictions from overlapping masks and obtain all instances concurrently.
When applied to large volumetric biological images, the resulting method proved to be strongly robust to noise and compared favorably to competing methods that need super-pixels and hence more hyper parameters.
The proposed method also endows its predictions with an uncertainty measure, depending on the consensus of the overlapping  \maskname masks. In future work, we plan to use these uncertainty measures to estimate the confidence of individual instances, which could help facilitate the subsequent proof-reading step still needed in neuron segmentation.

	\bibliographystyle{splncs03}
	\bibliography{patchEmbeddings}

	\newpage
	
\renewcommand{\thesection}{S\arabic{section}}
\renewcommand{\thetable}{S\arabic{table}}
\renewcommand{\thefigure}{S\arabic{figure}}

\section{Supplementary material}

\subsection{Graph neighborhood structure and output instance segmentation}
\begin{figure}[t]
\centering
        \includegraphics[width=\textwidth,trim=1.80in 1.4in 1.8in 1.50in,clip]{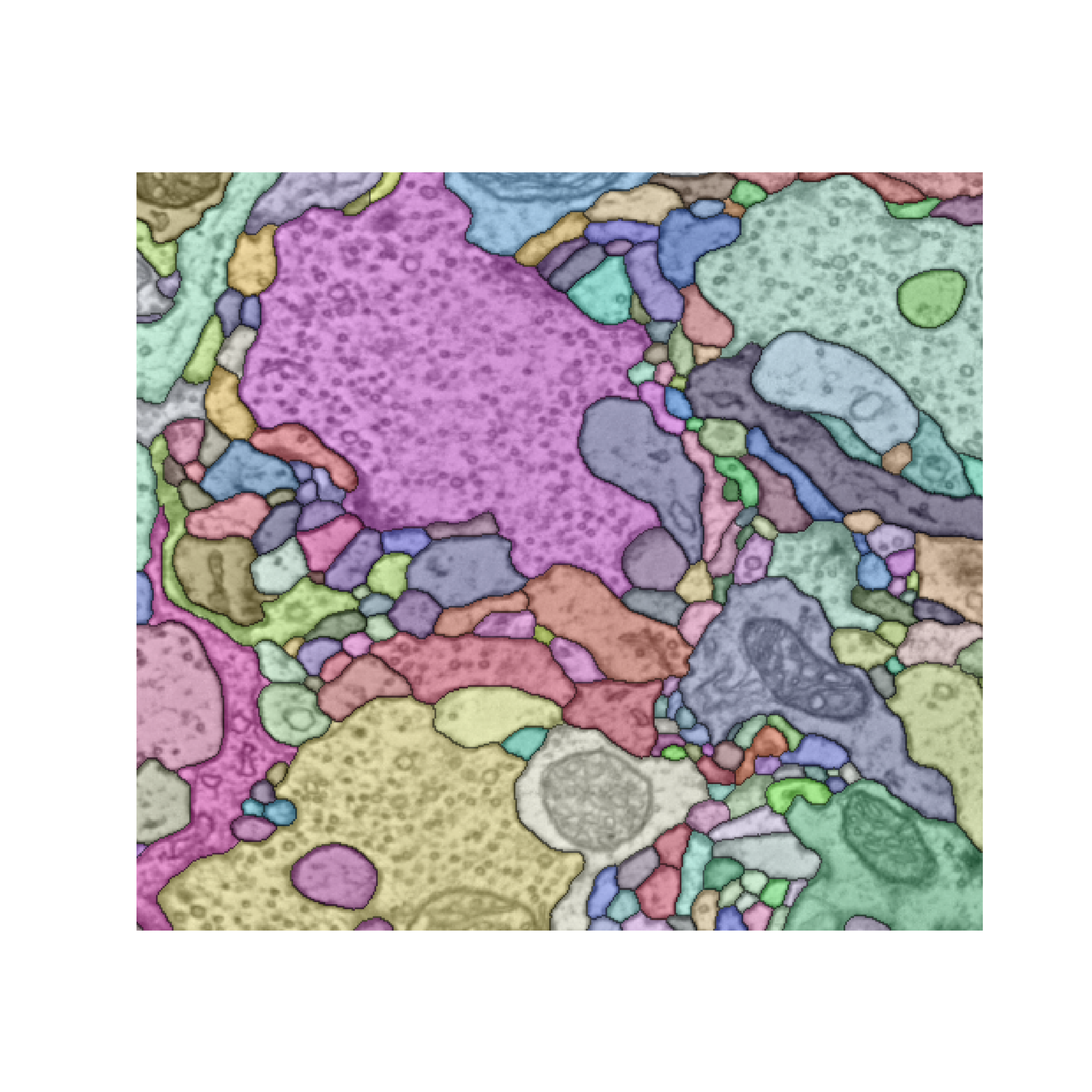} %
        \caption{Raw data from the validation set overlaid with the final instance segmentation obtained with our method: affinities are computed by averaging overlapping masks (MaskAggr); the final segmentation is achieved by running the Mutex Watershed algorithm on the obtained graph with positive and negative edge weights. Note that the data is 3D, hence the same color could be assigned to parts of segments that appear disconnected in 2D.}
    \label{fig:MWS_segm}
\end{figure}

In the first column of Table \ref{tab:neighborhood_structures}, we provide the neighborhood structure of the pixel grid-graph, which is very similar to the one used in related work \cite{wolf2018mutex,lee2017superhuman}.
In Fig. \ref{fig:MWS_segm}, we show the resulting instance segmentation obtained by computing affinities from \maskname masks and then running the Mutex Watershed algorithm on the obtained graph with positive and negative edge weights.

\paragraph{Removing small segments} -- After running the Mutex Watershed, we use a simple post-processing step to delete small segments on the boundaries, most of which are given by single-voxel clusters. On the neuron segmentation predictions, we deleted all regions with less than 200 voxels and used a seeded watershed algorithm to expand the bigger segments.

\subsection{Details on the model architecture}\label{sec:arch_details_suppl}
\begin{figure}[t]
\centering
        \includegraphics[width=\textwidth]{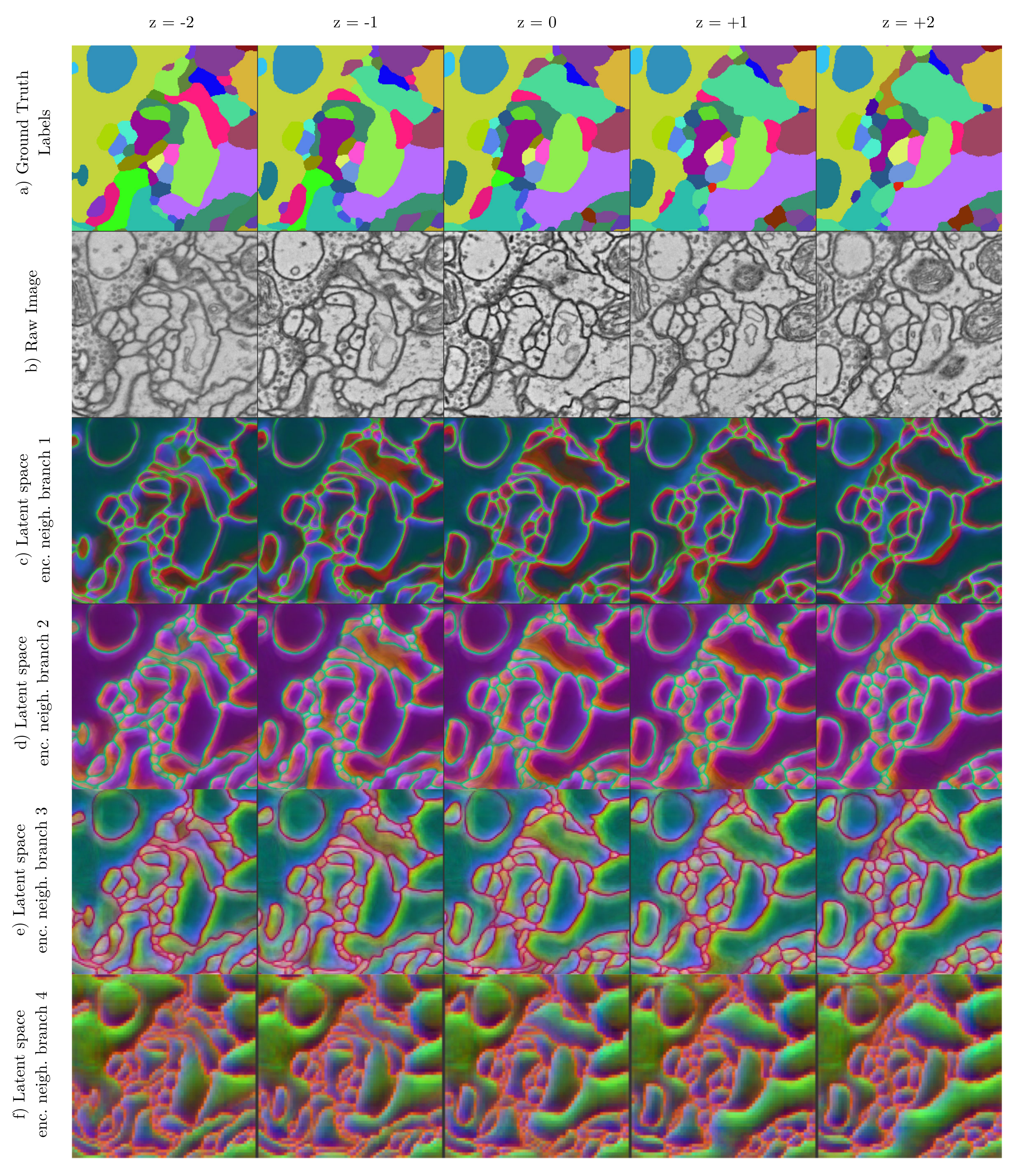} %
        \vspace{1em}
        \caption{\textbf{Visualization of the predicted single-instance mask latent spaces} -- Each column represents a 2D image from the 3D stack (only five are shown here). \textbf{(a)} Ground-truth labels. \textbf{(b)} Raw image given as input to the model. \textbf{(c-d-e-f)} Visualization of the first three principal components of the $32$-dimensional mask latent spaces predicted by the \encBr branches ENB$_{i=1,2,3,4}$ in our model. Note how latent spaces learned at different levels of the U-Net pyramid show different feature-scales, because they encode \maskname masks at different resolutions.}
    \label{fig:PCA_embeddings}
\end{figure}

Fig. \ref{fig:model_architecture} shows the details on the 3D-UNet architecture and Table \ref{tab:neighborhood_structures} lists the sparse neighborhood structures predicted by the \emph{\sparseBr branches}. Only the outputs at the highest resolution (given by branches SNB$_1$, ENB$_1$ and ENB$_2$) are used to compute edge weights in the pixel grid-graph. A visualization of the predicted single-instance mask latent spaces is given in Fig. \ref{fig:PCA_embeddings}.

The input volume has shape $272 \times 272\times12$ which is equivalent to a volume of $544\times 544\times 12$ voxels in the original resolution $4\times 4\times 40$ nm$^3$. Before to apply the loss, we crop the predictions to a shape $224\times 224\times 9$ in order to avoid border artifacts\footnote{Instead of cropping directly the final predictions, we perform several crops in the decoder part of the UNet model (see Upsample + Crop connections in Fig. \ref{fig:model_architecture}) in order to optimize GPU-memory usage.}.
The final model trained on all available ground truth labels is trained with a slightly larger input volume of $288\times 288\times 14$.

\subsection{Efficient Affinities for any Sparse Neighborhood}\label{sec:efficient_affs}
An advantage of training dense \maskname masks is that the graph $N$-neighborhood structure can be defined at prediction time, after the model has been trained. As an alternative to the method presented in Sec. \ref{sec:aggr_affs} that aggregates overlapping masks, here we propose the following efficient approach to predict affinities for a sparse neighborhood structure: Given a model that has been already trained end-to-end to predict encoded \maskname masks, we stack few additional convolutional layers that are trained to convert the $Q$-dimensional latent mask space to $N$ output feature maps representing affinities for the chosen sparse neighborhood structure. These last layers are not trained jointly with the full model, so in practice they are very quick and easy to train with a binary classification loss. By using this method, we avoid to decode all masks explicitly (one for each pixel) and achieve great time and memory savings.
As a result, we obtain a model that at inference time is no more memory-consuming than the current state-of-the-art approach predicting affinities only for a specific sparse neighborhood structure.

\subsection{Details on CREMI dataset and data augmentation}\label{sec:cremi_data_augm}
We test our method on the competitive CREMI 2016 EM Segmentation Challenge \cite{cremiChallenge} that is currently the neuron segmentation challenge with the largest amount of training data available. The dataset comes from serial section EM of \emph{Drosophila} fruit-fly brain tissue and consists of 6 volumes of $1250\times 1250\times 125$ voxels at resolution $4\times 4\times 40$ nm$^3$, three of which come with publicly available training ground truth. 
We achieved the best scores by downscaling the resolution of the EM data by a factor $(\frac{1}{2},\frac{1}{2},1)$, since this helped increasing the 3D context provided as input to the model.

\textbf{Data Augmentation} -- The data from the CREMI challenge is highly \linebreak anisotropic and contains artifacts like missing sections, staining precipitations and support film folds. 
To alleviate difficulties stemming from misalignment, we use a version of the data that was elastically realigned by the challenge organizers with the method of \cite{saalfeld2012elastic}.
In addition to the standard data augmentation techniques of random rotations, random flips and  elastic deformations, we simulate data artifacts.
In more detail, we randomly zero-out slices, introduce alignment jitter and paste artifacts extracted from the training data. Both \cite{funke2018large} and \cite{lee2017superhuman} have shown
that these kinds of augmentations can help to alleviate issues caused by EM-imaging artifacts. For zero-out slices, the model is trained to predict the ground-truth labels of the previous slice.
On the test samples, we run predictions for overlapping volumes and then average them.

\begin{figure}[t]
\centering
        \includegraphics[width=\textwidth]{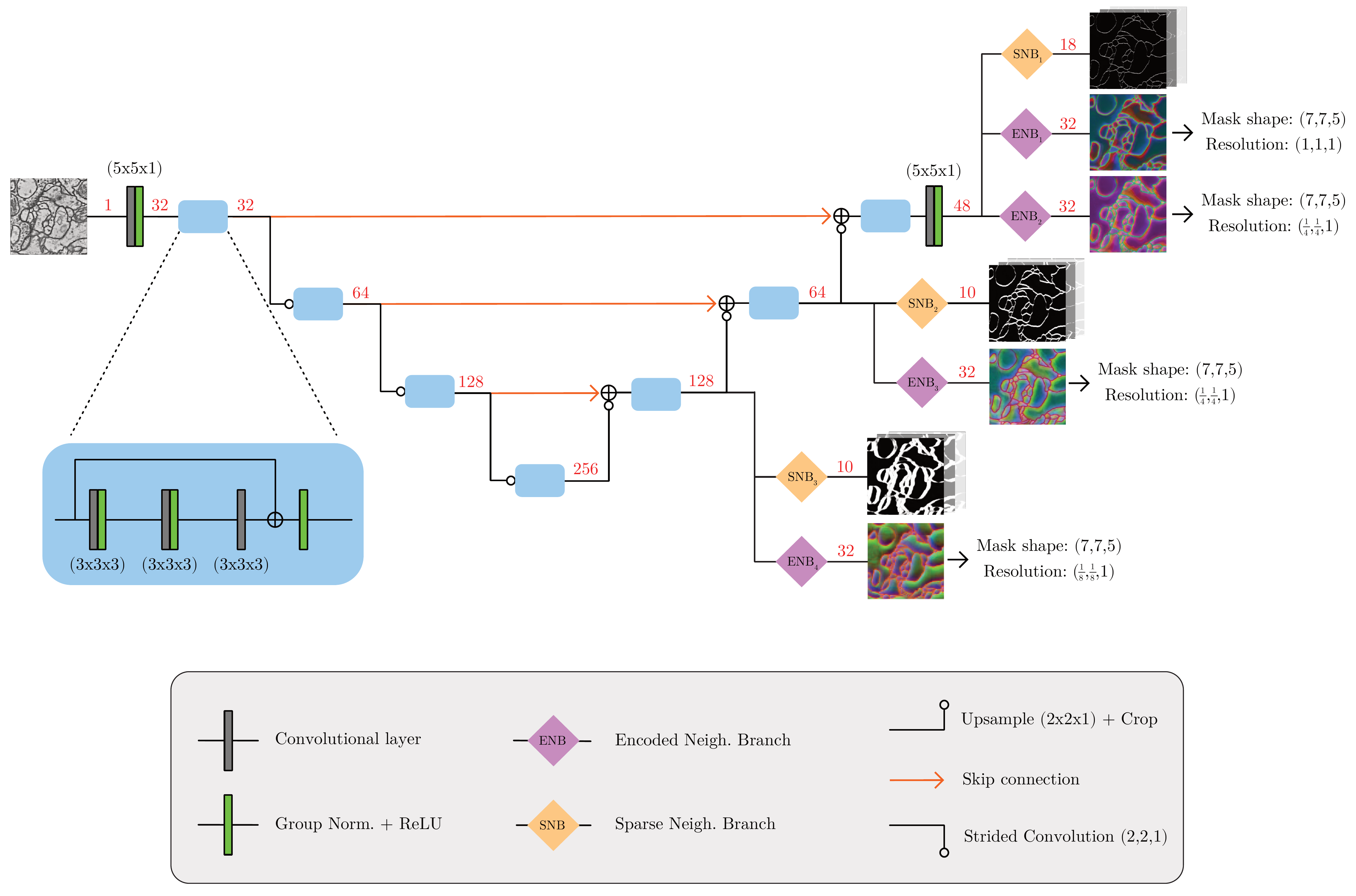} %
        \vspace{1em}
        \caption{\textbf{The architecture of the model}, which is strongly inspired by the 3D-UNet models proposed in \cite{lee2017superhuman,funke2018large}. 
        Red numbers indicate the number of used feature maps.
        As we explain in Sec. \ref{sec:models_details}, in this work we consider three models: i) a baseline model based on the three \emph{\sparseBr branches} SNB$_{i=1,2,3}$, shown in the figure; ii) another model based on the four \emph{\encBr branches} ENB$_{i=1,2,3,4}$; iii) and, finally, a combined model trained with all seven branches shown in the Figure.
        Even though the input of the model is a 3D volume, here, for simplicity, we show an example of 2D input image taken from the stack. As output of the \emph{\sparseBr branches} SNB$_{i=1,2,3}$, we show few channels representing some of the predicted affinities (see Table \ref{tab:neighborhood_structures} for details on the sparse neighborhood structures predicted by each branch SNB$_{i=1,2,3}$). We also show the first three principal components of the encoded masks predicted by the \emph{\encBr branches} ENB$_{i=1,2,3,4}$. All branches ENB$_{i=1,2,3,4}$ predict \maskname masks of the same window size $7 \times 7 \times 5$, but at different resolutions.}
    \label{fig:model_architecture}
\end{figure}

\begin{table}[t]
\small
\centering
        \begin{tabular}[t]{M{10em} M{8em} M{8em} M{8em} }
\thead{Graph neighborhood\\structure\\(16 neighbors)} & \thead{SNB$_1$\\(18 neighbors)} &  \thead{SNB$_2$\\(10 neighbors)}  & \thead{SNB$_3$\\(10 neighbors)} \\ \toprule 
(0, 0, -1)      & (0, 0, -1)    & (0, 0, -1)    & (0, 0, -1) \\
(-1, 0, 0)      & (-1, 0, 0)    & (-4, 0, 0)    & (-4, 0, 0) \\
(0, -1, 0)     & (0, -1, 0)     & (0, -4, 0)    & (0, -4, 0) \\
(-4, 0, 0)     & (-4, 0, 0)     & (0, 0, -2)    & (0, 0, -2) \\
(0, -4, 0)     & (0, -4, 0)     & (0, 0, -3)    & (0, 0, -3) \\
(-4, -4, 0)    & (-4, -4, 0)    & (0, 0, -4)    & (0, 0, -4) \\
(4, -4, 0)     & (4, -4, 0)     & (-14, 0, 0)   & (-12, 0, 0) \\
(-4, 0, -1)     & (-4, 0, -1)   & (0, -14, 0)   & (0, -12, 0) \\
(0, -4, -1)     & (0, -4, -1)   & (-14, -14, 0) & (-12, -12, 0) \\
(-4, -4, -1)    & (-4, -4, -1)  & (14, -14, 0)  & (12, -12, 0) \\
(4, -4, -1)     & (4, -4, -1)   &  - & - \\
(0, 0, -2)      & (0, 0, -2)    & - & - \\
(-8, -8, 0)    & (0, 0, -3)     & - & - \\
(8, -8, 0)     & (0, 0, -4)     & - & - \\
(-12, 0, 0)    & (-8, -8, 0)    & - & - \\
(0, -12, 0)    & (8, -8, 0)     & - & - \\
-                & (-12, 0, 0)   & - & - \\
-               & (0, -12, 0)   & - & - \\
        \end{tabular}
        \vspace{3em}
        \caption{\textbf{Sparse neighborhood structures} --  
        In this table, we represent sparse neighborhood structures (see for example the one shown in Fig. \ref{fig:main_figure}a) as lists of offsets $(\delta_x,\delta_y, \delta_z)$ indicating the relative coordinates of neighboring pixels with respect to the central pixel.
        The first column shows the neighborhood structure of the pixel grid-graph, such that each pixel / node is connected to 16 neighbors. In the following columns, we provide the neighborhood structures predicted by the three \emph{\sparseBr branches} SNB$_i$ used in our model (see architecture in Fig. \ref{fig:model_architecture}).
        These neighborhood structures were inspired by the ones used in \cite{wolf2018mutex,lee2017superhuman} but were adapted to our version of the CREMI data that is downscaled by a factor $(\frac{1}{2},\frac{1}{2},1)$. Note that the offsets provided here are given in the downscaled resolution.} \label{tab:neighborhood_structures}
\end{table}

\end{document}